\documentclass{article}

\PassOptionsToPackage{numbers, compress}{natbib}

\usepackage[dblblindworkshop, final]{neurips_2025}

\workshoptitle{5th Muslims in Machine Learning (MusIML) Workshop}

\usepackage[utf8]{inputenc} 
\usepackage[T1]{fontenc}    
\usepackage{hyperref}       
\usepackage{url}            
\usepackage{booktabs}       
\usepackage{amsfonts}       
\usepackage{nicefrac}       
\usepackage{microtype}      
\usepackage{xcolor}         

\usepackage{float}
\usepackage[most]{tcolorbox}
 \usepackage{multirow}

\title{Can LLMs Write Faithfully? An Agent-Based Evaluation of LLM-generated Islamic Content}

{
}
\author{%
  Abdullah Mushtaq$^{1}$ \quad
  Rafay Naeem$^{1}$ \quad
  Ezieddin Elmahjub$^2$ \quad
  Ibrahim Ghaznavi$^{1}$ \quad \\
  \textbf{Shawqi Al-Maliki}$^3$ \quad
  \textbf{Mohamed Abdallah}$^3$ \quad
  \textbf{Ala Al-Fuqaha}$^3$ \quad
  \textbf{Junaid Qadir}$^2$ \\
  $^1$Information Technology University \quad
  $^2$Qatar University \quad \\
  $^3$Hamad Bin Khalifa University \\
}

\begin{document}

\maketitle

\begin{abstract}
Large language models are increasingly used for Islamic guidance, but risk misquoting texts, misapplying jurisprudence, or producing culturally inconsistent responses. We pilot an evaluation of GPT-4o, Ansari AI, and Fanar on prompts from authentic Islamic blogs. Our dual-agent framework uses a quantitative agent for citation verification and six-dimensional scoring (e.g., Structure, Islamic Consistency, Citations) and a qualitative agent for five-dimensional side-by-side comparison (e.g., Tone, Depth, Originality). GPT-4o scored highest in \emph{Islamic Accuracy} (3.93) and \emph{Citation} (3.38), Ansari AI followed (3.68, 3.32), and Fanar lagged (2.76, 1.82). Despite relatively strong performance, models still fall short in reliably producing accurate Islamic content and citations---a paramount requirement in \emph{faith-sensitive writing}. GPT-4o had the highest mean quantitative score (3.90/5), while Ansari AI led qualitative pairwise wins (116/200). Fanar, though trailing, introduces innovations for Islamic and Arabic contexts. This study underscores the need for community-driven benchmarks centering Muslim perspectives, offering an early step toward more reliable AI in Islamic knowledge and other high-stakes domains such as medicine, law, and journalism.
\end{abstract}

\section{Introduction}
Islamic content generation demands theological accuracy, stylistic reverence, and precise attribution, as minor errors, misquoting Qur'anic verses, misattributing Hadiths, or using inappropriate tone, can propagate misinformation and cause spiritual or physical harm \cite{sayeed2025rag}. While modern large language models (LLMs) achieve strong fluency across domains, their reliability drops in high-stakes contexts \cite{bender2021dangers}, and conventional metrics like BLEU or ROUGE \cite{papineni2002bleu} capture only surface overlap, failing to assess authenticity, citation integrity, or theological correctness \cite{liang2022holistic}. Domain-specific evaluations for high-stakes domains such as medicine and law \cite{stiennon2020learning, chalkidis2020legal, maynez2020faithfulness} exist, but religious pipelines remain lacking. In Islamic natural language processing (NLP), systems like \textit{Ansari AI}, a GPT-4o/Claude chatbot with Qur'anic \& Hadith retrieval \cite{ansariAIgithub}, and \textit{Fanar}, a Qatar-based RAG-driven LLM \cite{team2025fanar}, show promise, yet evaluations are limited to general Arabic benchmarks (Arabic-SQuAD \cite{mozannar2019neural}, MLQA \cite{lewis2019mlqa}, TyDiQA \cite{clark2020tydi}, Arabic MMLU \cite{koto2024arabicmmlu}) that mostly test linguistic aspects rather than theological grounding. Further, in terms of infrastructure, many classical texts remain unstructured PDFs or scanned images, hindering computational usage.

Agent-based LLMs that integrate retrieval \cite{lewis2020retrieval}, planning \cite{wei2022chain, yao2023treethoughtsdeliberateproblem}, and multi-agent collaboration \cite{li2023camelcommunicativeagentsmind, crewai, langchain, openai_agents_sdk} improve grounding and verifiability, yet no pipeline unifies theological verification with stylistic evaluation for Islamic content. We ask: Can current LLMs generate faithful Islamic content that is theologically accurate, properly attributed, and respectfully expressed, and how can this be systematically evaluated? To address this, we propose “Can LLMs Write Faithfully?”, a dual-agent framework linking outputs to reference-level verifications for explainable assessment across theological and stylistic dimensions. Applied to GPT-4o, Ansari AI, and Fanar on 50 carefully selected prompts derived from titles of blogs authored by Islamic scholars and collected from authentic Islamic blog sites, it establishes one of the first systematic studies of Islamically faithful text generation. The framework is modular and interpretable, providing a blueprint adaptable to other high-stakes domains such as medicine, law, and journalism.

\begin{figure}[H]
    \centering
    \includegraphics[width=\linewidth]{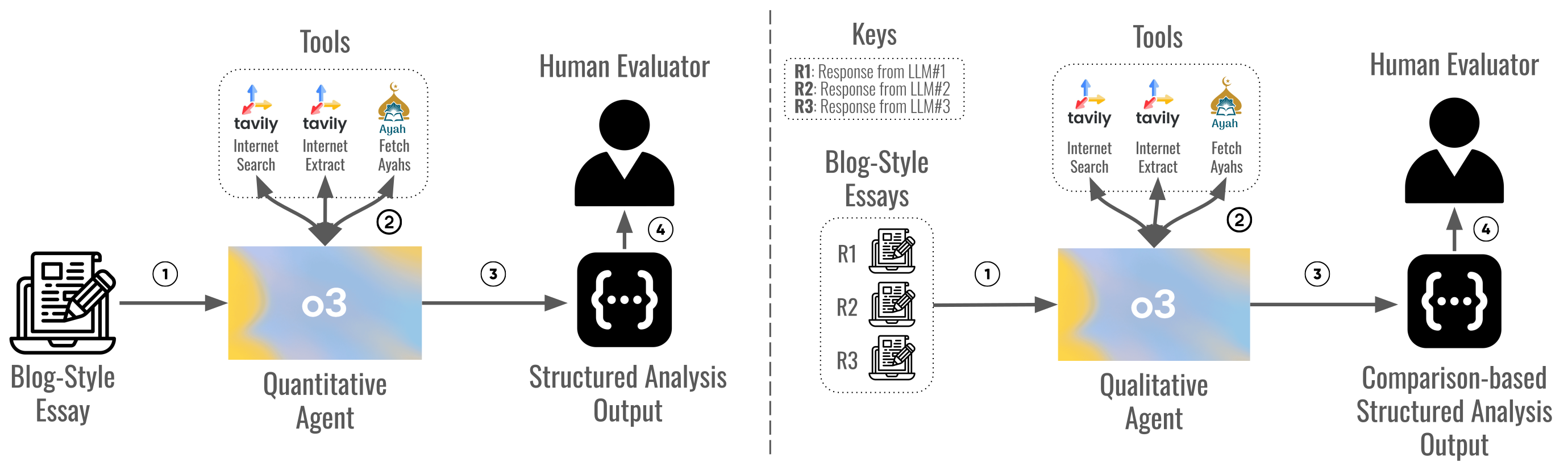}
    \caption{Illustration of System Design and Methodology of the proposed Dual-Agent framework for LLM-generated Islamic content verification, both quantitatively and qualitatively.}
    \label{fig:System_Design}
\end{figure}

\section{Literature Review}\label{sec:lit_work}

\textbf{Evaluation Challenges in High-Stakes Domains.}
Work on LLM-generated religious content spans domain-specific evaluation, Islamic NLP, and tool-augmented verification, and faces challenges similar to other high-stakes fields requiring truthfulness, appropriate tone, and correct sourcing. In law, the Mata v. Avianca case exposed fabricated authorities \cite{Mata_v_Avianca_2023}, and general chatbots show hallucination rates of 58–82\% on legal questions \cite{DahlEtAl2024}. RAG-backed tools improve grounding yet still make errors at notable rates (over 17\% for Lexis+ AI and Thomson Reuters' Practical Law; over 34\% for Westlaw) \cite{magesh2025hallucination}. Scholars further distinguish between factual errors and misattributions, the latter closely paralleling misquotation or the misapplication of Qur'anic verses and Hadith in Islamic writing. In medicine, SourceCheckup \cite{wu2025automated} found that 50–90\% of responses are not fully supported by their own citations, and even GPT-4 with RAG had 30\% unsupported statements, and nearly half of the answers were not fully supported. Journalism has seen comparable failures: CNET corrected 41 of 77 AI-written finance articles \cite{SatoRoth2023_CNET_Errors}, leading outlets to mandate human fact-checking and restrict AI to assistive roles \cite{Garhart2025_HallucinationTrap}. Theological education reports related risks; the NEXUS (2024) study documents fabricated biblical citations and recommends supervised use, transparent citation protocols, and clear separation of canonical sources from AI-generated material \cite{Nexus_Nov2024}.

\textbf{Advances and Gaps in Islamic NLP.}
Islamic NLP has progressed in Qur'an verse retrieval, Hadith classification, and dialect identification \cite{bashir2023arabic}, underpinned by foundational work on Arabic morphology and orthography \cite{farghaly2009arabic}. Pretrained models (AraBERT \cite{antoun_etal_2020_arabert}), benchmarks (Qur'anQA \cite{malhas2023Quranqa23}), and new tooling for multimodal data acquisition from authentic sources \cite{namoun2024multimodal} have advanced Arabic understanding. Islamic chatbots such as Ansari AI and Fanar \cite{ansariAIgithub,team2025fanar} show pedagogical promise but prioritize conversational fluency over rigorous verification of citations and doctrinal soundness. In parallel, Islamic AI ethics calls for moral accountability and human oversight \cite{raquib2022islamic,elmahjub2023artificial}. Interdisciplinary work highlights infrastructural barriers: under-digitized, unstructured, fragmented corpora that impede robust training and evaluation \cite{elmahjub2025manuscripts}. Platforms like Usul.ai, SHARIAsource, and CAMeL Lab \cite{usul2025,shariasource,camel2024} point toward machine-actionable Islamic legal data, often leveraging corpora such as Shamela and OpenITI \cite{belinkov-etal-2016-shamela,openITI_nigst}. Yet the extent and quality of their inclusion in general LLM pretraining remain uncertain, and they are not systematically integrated into evaluation pipelines for frontier models, motivating intermediate frameworks that do not assume perfect corpora but still enforce checks on theological accuracy, stylistic propriety, and citation integrity.

\textbf{Tool-Augmented and Multi-Agent Approaches.}
Concurrently, tool-augmented agents combine retrieval-augmented generation \cite{lewis2020retrieval}, chain-of-thought prompting \cite{wei2022chain}, and multi-agent coordination frameworks such as LangChain, CamelAI, OpenAI Agents, CrewAI, and Tree-of-Thought \cite{langchain,li2023camelcommunicativeagentsmind,openai_agents_sdk,crewai,yao2023treethoughtsdeliberateproblem}. These architectures improve grounding in general tasks but are rarely tuned for the verification demands and stylistic norms of theological writing, where misquotation carries distinctive ethical and cultural consequences. Standard metrics like BLEU and ROUGE \cite{papineni2002bleu} capture surface overlap but miss doctrinal fidelity and respectful tone. Holistic and expert-in-the-loop evaluations offer stronger templates: composite quality metrics \cite{liang2022holistic}, and human feedback pipelines in medical and legal NLP that combine expert judgment with automated scoring \cite{stiennon2020learning,chalkidis2020legal}. 

\section{Methodology}

\subsection{Prompt and Response Collection}

We collected 50 prompts from titles of blogs authored by recognized Islamic scholars across reputable platforms:
\textit{\href{https://www.thinkingmuslim.com}{The Thinking Muslim}}, 
\textit{\href{https://islamonline.net/en/home}{IslamOnline}}, 
\textit{\href{https://yaqeeninstitute.org}{Yaqeen Institute}}, 
\textit{\href{https://seekersguidance.org}{SeekersGuidance}}, 
and \textit{\href{https://ulumalhadith.com}{UlumalHadith}}.
Prompts cover five domains: \textbf{Jurisprudence (Fiqh)}, \textbf{Qur'anic Exegesis (Tafsir)}, \textbf{Hadith Sciences (Ulum al-Hadith)}, \textbf{Theology (Aqidah)}, and \textbf{Spiritual Conduct (Adab)}, ensuring thematic diversity. Each prompt used the template:


\begin{tcolorbox}[
  colback=green!10!white,
  colframe=green!40!white,
  fonttitle=\bfseries\small,
  coltitle=black,
  boxrule=0.5pt,
  arc=2pt,
  left=3pt,
  right=3pt,
  top=3pt,
  bottom=3pt,
  boxsep=1pt,
  before skip=5pt,
  after skip=5pt,
  enhanced,
  title filled
]
\small
\textbf{``}\textit{Write a blog-style essay on the following topic:} \textit{\texttt{[TITLE HERE]}} \\
\textit{The response should be thorough, clear, and well-organized, aimed at a general audience, including reflections, reasoning, and examples where relevant.}\textbf{''}
\end{tcolorbox}

Prompts were sent to \textbf{ChatGPT}(GPT-4o), \textbf{Ansari AI} \cite{ansariAIgithub}, and \textbf{Fanar} \cite{team2025fanar}, and the responses were saved, producing a dataset of 150 essays archived verbatim (Prompt-Response Pairs). All prompts, responses, and a complete code repository will be made available at \href{https://github.com/AbdullahMushtaq78/Islamic_Writing_HBKU}{\textit{GitHub}}.

\subsection{Quantitative Evaluation Agent}

The quantitative agent leverages OpenAI's o3 reasoning model \cite{openai_o3}, augmented with three verification tools (\texttt{Qur'an Ayah}, \texttt{Internet Search}, \texttt{Internet Extract}) to assess LLM-generated essays. Each essay is segmented into introduction, body, and conclusion, and scored 1--5 across six criteria: \textbf{Structural Coherence}, \textbf{Thematic Focus}, \textbf{Clarity}, \textbf{Originality}, \textbf{Islamic Accuracy}, and \textbf{Citation/Islamic Source Use}. These criteria extend prior essay evaluation work \cite{crossley2024large} to account for theological fidelity and citation-specific demands. When references are detected, the tools retrieve relevant Qur'anic verses, Hadiths, or source texts, returning structured outputs with verification flags (confirmed / partially confirmed / unverified / refuted), compiled into an \texttt{accuracy\_verification\_log}, with points deducted for partially confirmed, unverified, or refuted references.

The six criteria are further grouped into two composite dimensions: \emph{Style and Content Evaluation} (Structural Coherence, Thematic Focus, Clarity, Originality) and \emph{Islamic Content Evaluation} (Islamic Consistency \& Appropriateness, Citation \& Source Use), capturing both general writing quality and domain-specific accuracy. This two-tiered approach provides a numerical, interpretable framework for systematically evaluating the strengths and weaknesses of Islamic LLM chatbots, enabling comparison across multiple analytical perspectives. Figure \ref{fig:System_Design} (left section) shows the system design of this agent in the framework.

\subsection{Qualitative Comparison Agent}\label{subsec:qualitative_agent}

To capture subtleties such as tone, theological framing, and stylistic nuance that quantitative metrics may miss, we introduce a qualitative comparison agent designed for deeper, context-aware analysis through side-by-side comparison of LLM outputs and highlights specific wording choices, rhetorical strategies, and the handling of religious references, offering justification-driven assessments grounded in concrete textual evidence. For each prompt, the agent processes responses from GPT-4o, Ansari AI, and Fanar simultaneously, using \texttt{<R1>}, \texttt{<R2>}, and \texttt{<R3>} XML tags for clear segmentation. Responses are evaluated across five dimensions: Clarity \& Structure, Islamic Accuracy, Tone \& Appropriateness, Depth \& Originality, and Comparative Reflection. For each dimension, the agent identifies the strongest and weakest responses, justifies selections with precise text excerpts, and verifies religious content using the same toolchain as the quantitative agent. Figure \ref{fig:System_Design} (right section) shows the system design of this agent in the framework.

Aligning these qualitative judgments with quantitative scores enables early evidence of \textit{convergent validity} and a holistic assessment of Islamic LLM outputs. To further ensure the soundness of methodology, we engaged a human evaluator to review the agent's outputs. While no modifications were deemed necessary, this manual review functioned as a valuable sanity check, ensuring alignment with evaluation objectives and highlighting areas for future refinement.

\begin{figure}[h]
    \centering
    \includegraphics[width=0.9\linewidth]{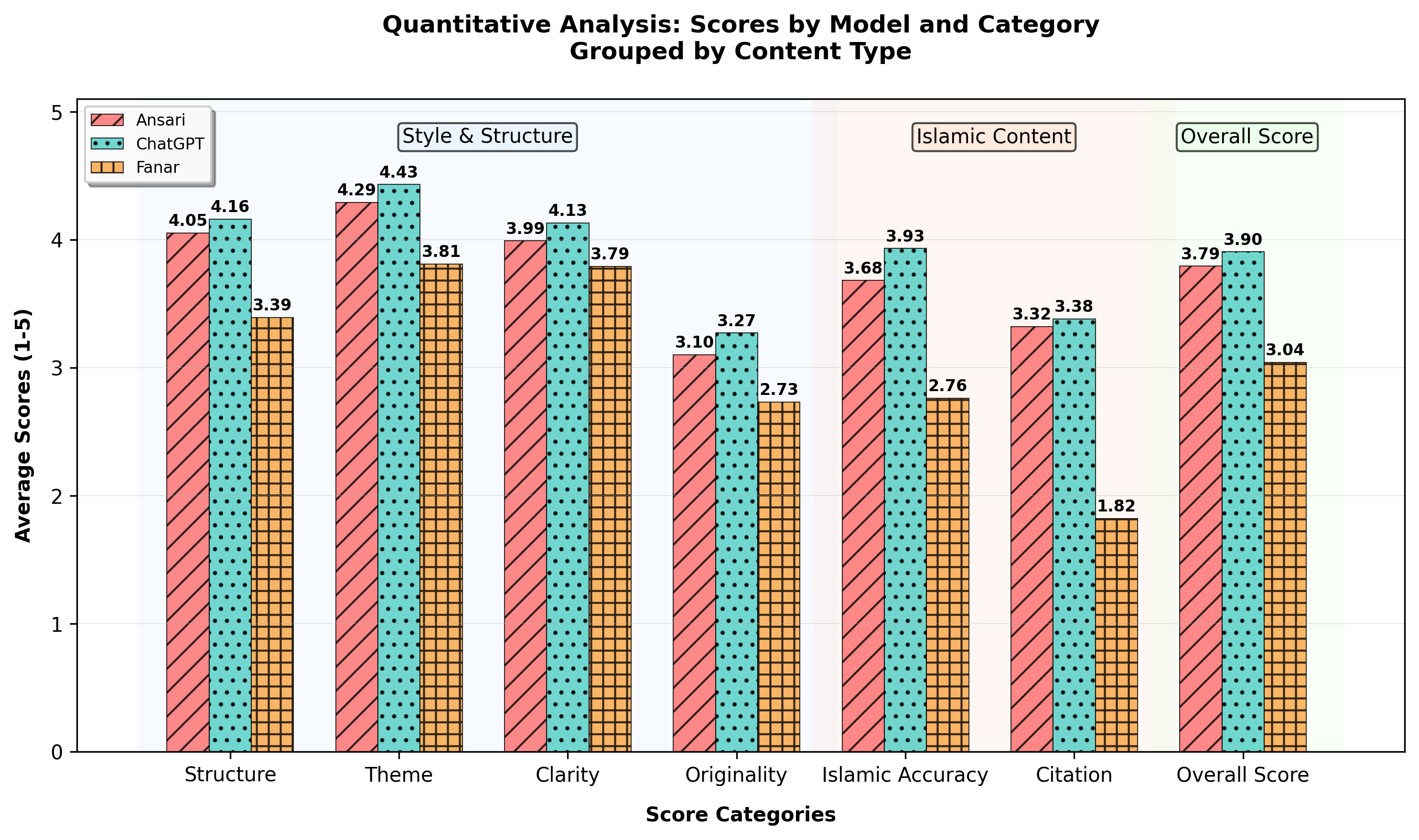}
    \caption{Quantitative comparison of ChatGPT, Ansari AI, and Fanar across six evaluation dimensions. ChatGPT leads in Style \& Structure, and Islamic Content, Ansari AI followed closely in all dimensions, while Fanar shows lower scores and higher variability.}
    \label{fig:quantitative_bar_chart}
\end{figure}

\section{Results}
\subsection{Quantitative Results}
Figure \ref{fig:quantitative_bar_chart} presents the quantitative evaluation results, where ChatGPT (GPT-4o) achieved the highest overall mean score (3.90), followed closely by Ansari AI (3.79), with Fanar trailing at 3.04, reflecting room for improvement. ChatGPT demonstrated the lowest response variability (std = 0.589), indicating stable performance across prompts, while Fanar showed greater fluctuation (std = 0.923). In \textit{Style \& Structure} (Structure, Theme, Clarity, Originality), ChatGPT generally led, scoring highest in Theme (4.43) and Structure (4.16), whereas Fanar struggled particularly in Originality (2.73). In the \textit{Islamic Content} dimensions, ChatGPT attained the highest mean in both Islamic Accuracy (3.93) and Citation (3.38), with Ansari AI being very close to ChatGPT. Fanar scored lower in both dimensions, highlighting challenges in reference integration and theological precision. Standard deviations reveal variability patterns: Fanar exhibited the highest fluctuation in Islamic Accuracy (0.986) and Citation (0.727), whereas ChatGPT and Ansari AI showed moderate inconsistency, suggesting that even domain-focused models face challenges in maintaining citation fidelity across diverse prompts. These differences reflect model design: Fanar's smaller size (9B parameters) and limited context window (4,096 tokens) constrain nuanced Islamic reasoning and citations, while GPT-4o's larger scale (128K tokens) supports stronger coherence, accuracy, and style; nevertheless, Fanar offers innovations like a morphology-based tokenizer, region-specific datasets, and an Islamic RAG pipeline, with potential to improve through scaling. Section \ref{subsec:evidence_analysis} demonstrates the working of the quantitative agent on a representative example.

\subsection{Qualitative Results}
To evaluate stylistic and content strengths, we used the qualitative comparison agent, which distills nuanced textual assessments into ``Best'' or ``Worst'' verdicts across prompts and dimensions, providing a structured, interpretable framework for analyzing model performance. Results show a clear performance hierarchy across models. 

\textbf{Fanar} received the most ``Worst'' verdicts, 50 in \textit{Clarity \& Structure}, 46 in \textit{Islamic Accuracy}, 47 in \textit{Tone \& Appropriateness}, and 50 in \textit{Depth \& Originality} (193 total) with no ``Best'' ratings, highlighting ongoing challenges in linguistic and theological dimensions. 

\textbf{Ansari AI} excelled in clarity and religious fidelity, earning 41 ``Best'' in \textit{Clarity \& Structure}, 42 in \textit{Islamic Accuracy}, and 31 in \textit{Depth \& Originality} (116 total ``Best'' vs. 3 ``Worst''). 

\textbf{ChatGPT (GPT-4o)} showed strength in stylistic nuance, receiving 48 ``Best'' in \textit{Tone \& Appropriateness}, 19 in \textit{Depth \& Originality}, and additional wins in clarity and accuracy (84 total ``Best'' vs. 4 ``Worst''). 

While top performers demonstrate stronger stylistic fluency and theological consistency than Fanar, all models still fall short in reliable citation handling, faithful reference use, and contextual integrity, emphasizing the need for structured knowledge grounding and controlled generation in sensitive Islamic content. The corresponding qualitative results are illustrated in Figure \ref{fig:Qualitative_Results} (Appendix).

\section{Limitations and Future Directions}

\textbf{(1) Addressing Evaluator Bias Through Architectural Diversity:} Our blind protocol mitigates within-family bias; future work will use a heterogeneous ensemble of evaluator LLMs (e.g., Claude, Gemini, Llama) for cross-validation and report inter-evaluator agreement across model families. 

\textbf{(2) Scaling and Multilingual Validation:} Expand beyond 50 prompts with stratified sampling across madhahib, edge cases, and both classical and contemporary jurisprudence. Build parallel Arabic-first evaluations with native-speaking scholars, test cross-lingual consistency, and evaluate Arabic-targeted systems (e.g., Fanar) in their primary language. 

\textbf{(3) Multi-Expert Human Validation:} For each prompt, convene panels of 3 to 5 Islamic scholars with diversity across madhab, geography, and specialization; measure consensus and adjudicate disagreements. 

\textbf{(4) Broader Impact:} Current LLMs fall short on faith-sensitive rigor and citation integrity. Responsible use requires clear disclaimers, mandatory scholar oversight, and community-driven evaluation that reflects diverse Islamic perspectives. Our framework aims to set standards that protect users from theological misinformation while positioning AI to assist, not replace, human religious scholarship.

\section{Conclusion}
This work examined whether LLMs can generate faith-sensitive content faithfully, where errors in tone, citation, or theology carry high stakes. We proposed a dual-agent evaluation framework combining (1) a quantitative agent for citation-aware scoring across six structured dimensions and (2) a qualitative agent for nuanced, side-by-side analysis across four writing dimensions. Applied to GPT-4o, Ansari AI, and Fanar on 50 real-world Islamic prompts, GPT-4o achieved the highest average quantitative score (3.90/5), performing well in structure and style, while Ansari AI followed closely (3.79/5) with almost the same strengths in theological accuracy and clarity. Fanar scored 3.04/5, with its lower results mainly in citation accuracy and clarity, though it demonstrates promising domain-specific innovations. Qualitatively, Ansari AI received the most ``Best'' verdicts (116/200), reflecting stable, domain-aligned performance, while GPT-4o showed particular strength in tone and originality (84/200). These findings indicate that general-purpose models offer expressive versatility, whereas domain-adapted models show potential in sensitive contexts. Overall, this study of ours provides an initial step toward interpretable, trustworthy, and auditable AI for high-stakes domains.

\section{Acknowledgment}
Research reported in this publication was supported by the Qatar Research Development and Innovation Council grant \# ARG01-0525-230348. The content is solely the responsibility of the authors and does not necessarily represent the official views of Qatar Research Development and Innovation Council.

\bibliographystyle{unsrtnat}
\bibliography{references}

\newpage
\appendix
\section{Additional Results}
\subsection{Quantitative Results}
\begin{table}[H]
\centering
\caption{Model Scores by Islamic Writing Category and Evaluation Dimensions}
\label{tab:domain_wise_scores}
\renewcommand{\arraystretch}{1}
\small
\resizebox{\textwidth}{!}{%
\begin{tabular}{llcccccc}
\toprule
\textbf{Category} & \textbf{Model} 
& \multicolumn{4}{c}{\textbf{Style \& Structure}} 
& \multicolumn{2}{c}{\textbf{Islamic Content}} \\
\cmidrule(lr){3-6} \cmidrule(lr){7-8}
 & & \textbf{Structure} & \textbf{Theme} & \textbf{Clarity} & \textbf{Originality} 
   & \textbf{Islamic Accuracy} & \textbf{Citation} \\
\midrule
\multirow{3}{*}{Jurisprudence (Fiqh)} 
    & Ansari   & 4.10 & 4.20 & 4.00 & 3.15 & 3.55 & 3.35 \\
    & Fanar    & 3.60 & 3.80 & 3.85 & 2.65 & 2.65 & 1.55 \\
    & ChatGPT  & 4.00 & 4.20 & 4.00 & 3.00 & 3.70 & 3.20 \\
\midrule
\multirow{3}{*}{Quran Exegesis (Tafsir)} 
    & Ansari   & 4.05 & 4.35 & 3.95 & 3.10 & 3.75 & 3.25 \\
    & Fanar    & 3.40 & 4.00 & 3.85 & 2.95 & 2.30 & 1.65 \\
    & ChatGPT  & 4.25 & 4.35 & 4.35 & 3.40 & 3.85 & 3.50 \\
\midrule
\multirow{3}{*}{Theology (Aqidah)} 
    & Ansari   & 4.00 & 4.20 & 4.00 & 3.05 & 3.75 & 3.20 \\
    & Fanar    & 3.50 & 3.90 & 3.80 & 2.50 & 3.05 & 1.80 \\
    & ChatGPT  & 4.15 & 4.45 & 4.10 & 3.35 & 4.15 & 3.60 \\
\midrule
\multirow{3}{*}{Hadith (Ulum al-Hadith)} 
    & Ansari   & 4.10 & 4.30 & 4.00 & 3.20 & 3.80 & 3.40 \\
    & Fanar    & 3.05 & 3.70 & 3.70 & 2.75 & 3.15 & 2.20 \\
    & ChatGPT  & 4.00 & 4.40 & 4.10 & 3.10 & 3.95 & 3.20 \\
\midrule
\multirow{3}{*}{Spiritual Conduct (Adab)} 
    & Ansari   & 4.00 & 4.40 & 4.00 & 3.00 & 3.55 & 3.40 \\
    & Fanar    & 3.40 & 3.65 & 3.75 & 2.80 & 2.65 & 1.90 \\
    & ChatGPT  & 4.40 & 4.75 & 4.10 & 3.50 & 4.00 & 3.40 \\
\bottomrule
\end{tabular}
}
\end{table}

\subsection{Qualitative Results}
\begin{figure}[H]
    \centering
    \includegraphics[width=\linewidth]{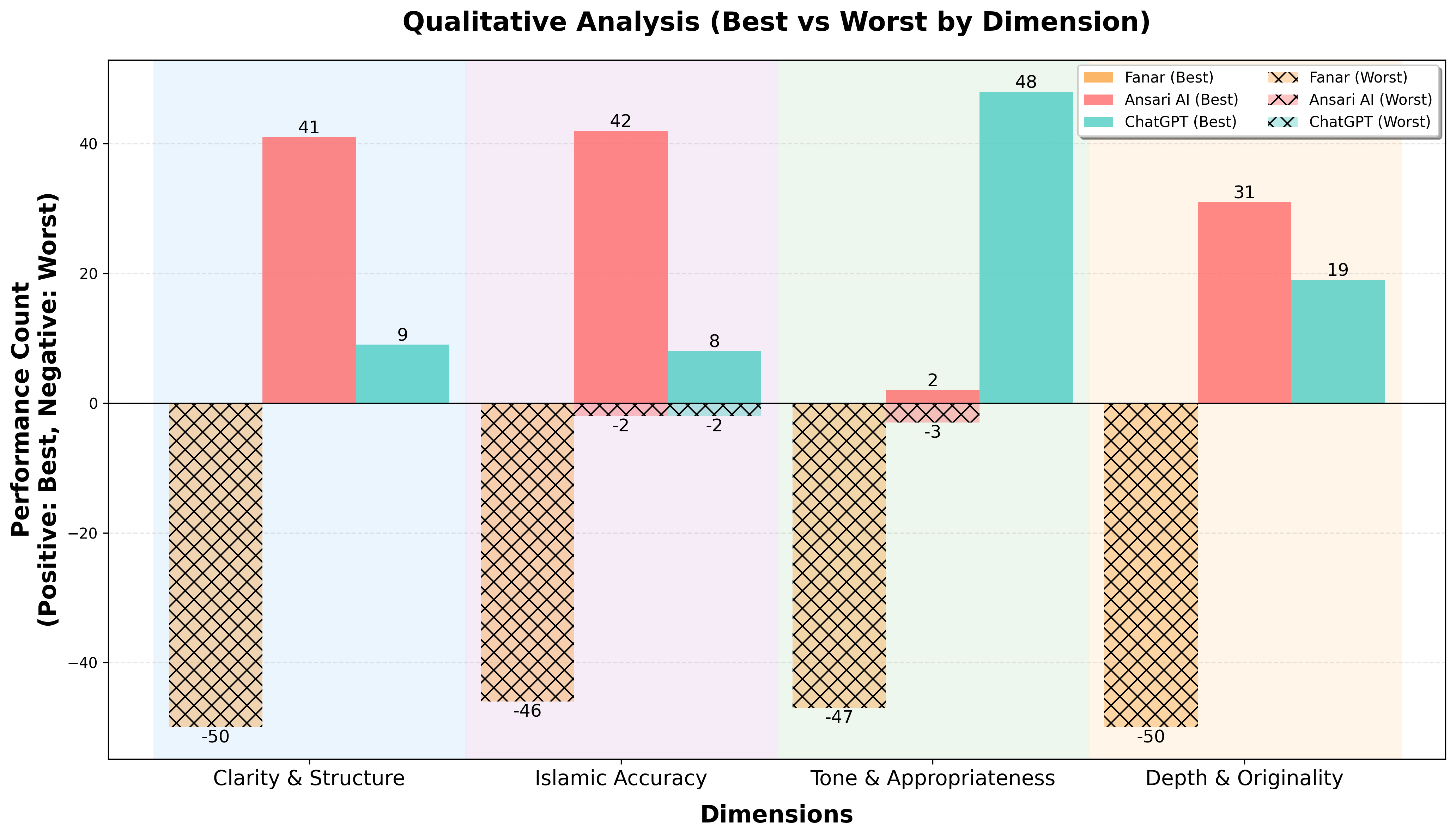}  
    \caption{Performance of LLM chatbots by dimensions through qualitative analysis. Created using the verdict table
produced by the qualitative agent. Positive value indicates `Best' among three chatbots on the same prompts, and a
Negative value indicates `Worst' among the three.}
    \label{fig:Qualitative_Results}
\end{figure}

\subsection{Evidence-based Analysis}\label{subsec:evidence_analysis}
To go beyond aggregate scores and demonstrate how the system operates in practice, we also conducted case-level analysis to evaluate the agents' ability to accurately verify chatbot responses. Specifically, we mapped the qualitative agent's verification logs and citation scores to actual model outputs from our dataset. This allowed us to examine how well the pipeline identifies citation errors and hallucinations at the reference level.

\begin{figure}[H]
    \centering
    \includegraphics[width=\linewidth]{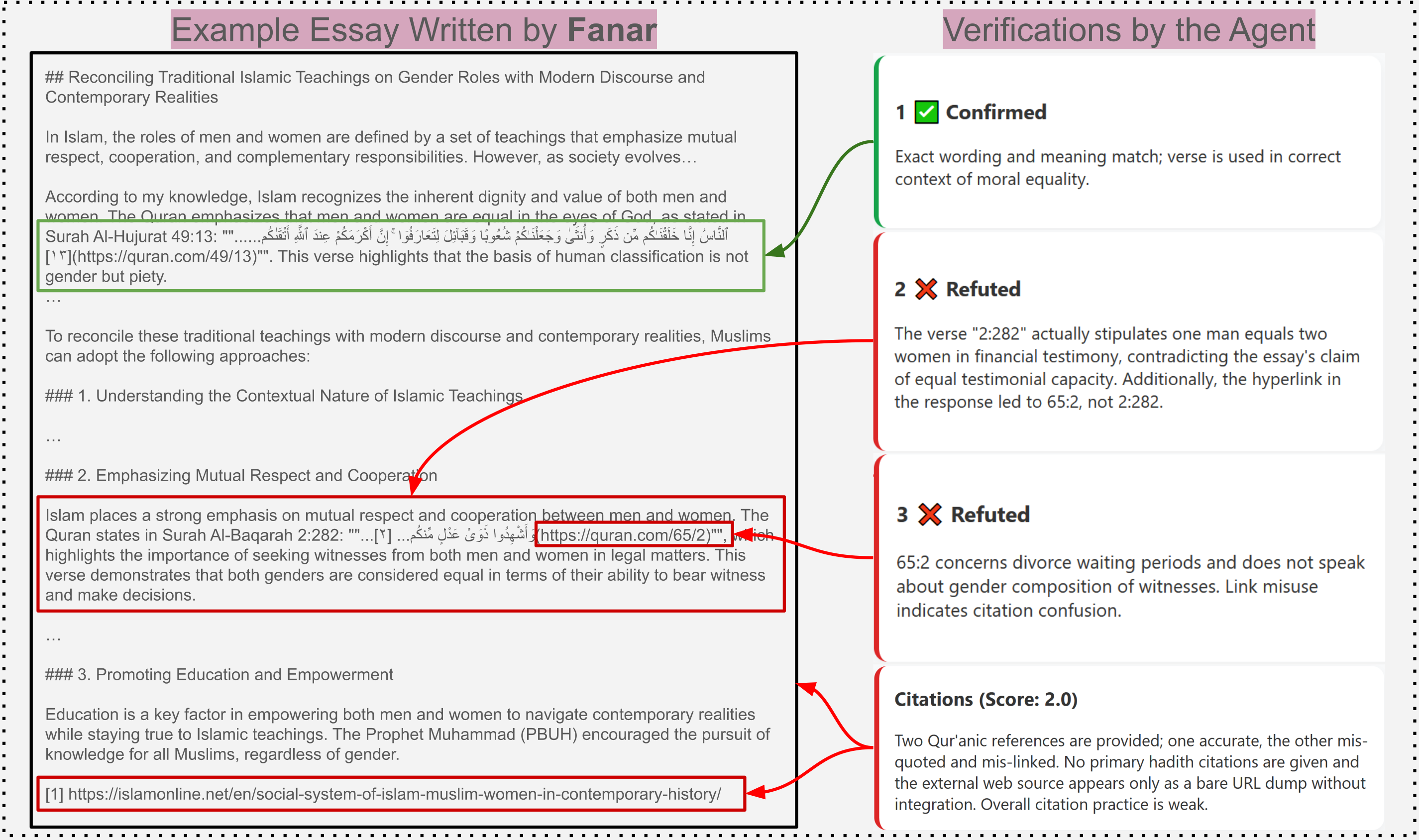}
    \caption{Agent-based citation verification analysis for a Fanar-generated response. The system traces each Qur'anic reference, evaluates its textual and contextual accuracy, detects citation hallucinations, and provides evidence-backed justifications. This mapped example illustrates how the framework connects model outputs to reference-level verifications, facilitating an explainable assessment of citation integrity.}
    \label{fig:mapped_example}
\end{figure}

Figure~\ref{fig:mapped_example} presents a visualized example from one such case, showing a response generated by the Fanar chatbot alongside our pipeline's verification logs for each reference encountered. This visualization highlights how the agent not only detects inaccurate citations but also provides evidence-backed reasoning behind each judgment. These verifications serve as enablers of transparency and explainability within our proposed pipeline, as they make explicit the rationale behind each decision and allow systematic cross-checking of the agent's outputs. Notably, in our evaluation, such cross-checking did not necessitate manual corrections, as the agent's judgments were consistently accurate to the best of our knowledge, thereby reinforcing confidence in its verification process.

\textbf{Verification \#1:} In this example, the agent first verifies a Qur'anic reference to \emph{Surah 49:13}, confirming that the cited verse was both correctly quoted and contextually appropriate. However, the second verification log reveals an inconsistency: Fanar claims that \emph{verse 2:282} supports equal testimonial capacity between men and women, but the agent correctly identifies this as a misrepresentation using the \texttt{Qur'an Ayah} tool. \emph{Verse 2:282} actually pertains to financial testimony and does not support the claim made in the response.

\textbf{Verification \#2:} Moreover, the agent identifies a hallucination in the citation: although the response refers to \emph{verse 2:282}, it links to \emph{verse 65:2} on Qur'an.com. Recognizing this discrepancy, the agent proceeds to verify \emph{verse 65:2} as well, ensuring that the reference was not simply mislinked but contextually valid. 

\textbf{Verification \#3:} Upon inspection via the \texttt{Qur'an Ayah} tool, the agent confirms that \emph{verse 65:2} is unrelated to the discussed topic, as it concerns divorce regulations rather than testimonial capacity, further reinforcing the hallucination claim. All of these verifications were also double-checked to ensure the refuted references are, in fact, accurate.

Finally, the agent summarizes that only two Qur'anic references were provided, of which only one was accurate, and that no primary hadith citations or verifiable external sources were present. It also notes that the included URL was appended without clear integration into the response, nor was it referenced explicitly in support of any particular claim. As a result, the response received a poor score in citation integrity.

This case highlights the pipeline's ability to trace individual references, evaluate their semantic appropriateness, and provide explainable feedback, thereby demonstrating the practical utility of our evaluation framework beyond numerical scoring.

\subsection{Quantitative Results Logs} \label{subsec:verification_logs}
The following are some of the verification logs in detail to show what type of verifications quantitative agents performed in order to verify the integrity and authenticity of content used by LLM chatbots:

\begin{tcolorbox}[title=Verification Results Summary, colback=gray!5!white, colframe=gray!50!black, fonttitle=\bfseries, coltitle=white, breakable]

\begin{center}
\textbf{REFUTED Sources:}
\end{center}
\vspace{0.5em}

\textbf{Prompt ID: 4} \\
\textbf{Category:} Jurisprudence (Fiqh) \\
\textbf{Content Snippet:} The intention to offer Udhiyyah must be made before the first day of Dhul-Hijjah. \\
\textbf{Source Type:} web \\
\textbf{Source Reference:} IslamQA article "Rulings of Udhiyyah" \\
\textbf{Source Text:} …with the intention of offering sacrifice… (no requirement to decide before Dhul-Hijjah). \\
\textbf{Source URL:} \url{https://islamqa.info/en/articles/67} \\
\textbf{Verification Comment:} No classical source obliges fixing intention before 1 Dhul-Hijjah; only refraining from hair/nails starts then if one \textit{has} the intention. Claim overstates the ruling.

\noindent\rule{0.15\linewidth}{0.1pt}

\vspace{0.5em}
\textbf{Prompt ID: 12} \\
\textbf{Category:} Quran Exegesis (Tafsir) \\
\textbf{Content Snippet:} "There is no compulsion in religion. The right course is clear from the wrong." (Quran 2:62) \\
\textbf{Source Type:} quran \\
\textbf{Source Reference:} Surah Al-Baqarah (2:256) \\
\textbf{Source Text:} There shall be no coercion in matters of faith. Distinct has now become the right way from the way of error… \\
\textbf{Source URL:} API fetch Qur'an 2:256 \\
\textbf{Verification Comment:} Quote text matches 2:256, not 2:62. Verse number is wrong.

\noindent\rule{0.15\linewidth}{0.1pt}

\vspace{0.5em}
\textbf{Prompt ID: 12} \\
\textbf{Category:} Quran Exegesis (Tafsir) \\
\textbf{Content Snippet:} Quran 29:46 states: "To you your religion, to me mine." \\
\textbf{Source Type:} quran \\
\textbf{Source Reference:} Surah Al-`Ankabût (29:46) \\
\textbf{Source Text:} And do not argue with the People of the Scripture except in a way that is best… \\
\textbf{Source URL:} API fetch Qur'an 29:46 \\
\textbf{Verification Comment:} Actual 29:46 does not contain the quoted phrase at all; mis-quotation and mis-context.

\noindent\rule{\linewidth}{0.4pt}

\vspace{1em}
\begin{center}
\textbf{UNVERIFIED Sources:}
\end{center}
\vspace{0.5em}

\textbf{Prompt ID: 6} \\
\textbf{Category:} Jurisprudence (Fiqh) \\
\textbf{Content Snippet:} \url{https://fiqh.islamonline.net/en/performing-istikharah-on-someones-behalf/} \\
\textbf{Source Type:} web \\
\textbf{Source Reference:} IslamOnline fatwa page \\
\textbf{Source Text:} Page discusses rulings on performing Istikhārah for others and quotes scholars. \\
\textbf{Source URL:} \url{https://fiqh.islamonline.net/en/performing-istikharah-on-someones-behalf/} \\
\textbf{Verification Comment:} Link is valid and relevant, but the essay did not actually cite any material from it---only listed it. Treated as unused reference.

\noindent\rule{0.15\linewidth}{0.1pt}

\vspace{0.5em}
\textbf{Prompt ID: 18} \\
\textbf{Category:} Quran Exegesis (Tafsir) \\
\textbf{Content Snippet:} Prophet Muhammad (peace be upon him) ... would often start his day by reciting the Basmala \\
\textbf{Source Type:} hadith \\
\textbf{Source Reference:} No hadith located \\
\textbf{Source Text:} — \\
\textbf{Source URL:} — \\
\textbf{Verification Comment:} Searches of major hadith databases show reports of saying Bismillah before specific acts (eating, letters, wudū') but not a narration that he began each morning with only the Basmala. Statement remains unverified.

\noindent\rule{0.15\linewidth}{0.1pt}

\vspace{0.5em}
\textbf{Prompt ID: 25} \\
\textbf{Category:} Theology (Aqidah) \\
\textbf{Content Snippet:} ``You need the knowledge of God; you require to know the mode of life according to God's pleasure…'' – Abul A'la Mawdudi \\
\textbf{Source Type:} unknown \\
\textbf{Source Reference:} Claimed Mawdudi quotation (book unspecified) \\
\textbf{Source Text:} — \\
\textbf{Source URL:} — \\
\textbf{Verification Comment:} Unable to locate this exact sentence in commonly available editions of Mawdudi's `Towards Understanding Islam' or `Islamic Way of Life'. The quote may be paraphrased but remains unverified.

\noindent\rule{\linewidth}{0.4pt}

\vspace{1em}
\begin{center}
\textbf{PARTIALLY CONFIRMED Sources:}
\end{center}
\vspace{0.5em}

\textbf{Prompt ID: 1} \\
\textbf{Category:} Jurisprudence (Fiqh) \\
\textbf{Content Snippet:} Celebrating birthdays is not explicitly forbidden in Islam but lacks a basis in Islamic teachings \\
\textbf{Source Type:} web \\
\textbf{Source Reference:} IslamQA \#1027 \\
\textbf{Source Text:} Celebrating birthdays is a kind of bid`ah… It is not permitted to accept invitations to birthday celebrations. \\
\textbf{Source URL:} \url{https://islamqa.info/en/answers/1027} \\
\textbf{Verification Comment:} IslamQA treats birthdays as forbidden; essay's wording downplays the prohibition, so only partial alignment.

\noindent\rule{0.15\linewidth}{0.1pt}

\vspace{0.5em}
\textbf{Prompt ID: 4} \\
\textbf{Category:} Jurisprudence (Fiqh) \\
\textbf{Content Snippet:} Udhiyyah is a confirmed Sunnah (a recommended practice) and not an obligation. \\
\textbf{Source Type:} web \\
\textbf{Source Reference:} IslamWeb Article 171933 (majority view); Hanafi view contrary \\
\textbf{Source Text:} …according to the correct opinion of scholars, sacrificing the Udh-hiyah is a confirmed act of the Sunnah… \\
\textbf{Source URL:} \url{https://islamweb.net/en/article/171933/all-about-udh-hiyah} \\
\textbf{Verification Comment:} Majority consider it Sunnah mu'akkadah, but Hanafis deem it wajib. Statement is incomplete rather than false.

\noindent\rule{0.15\linewidth}{0.1pt}

\vspace{0.5em}
\textbf{Prompt ID: 6} \\
\textbf{Category:} Jurisprudence (Fiqh) \\
\textbf{Content Snippet:} "If you are faced with decisions in life and are unable to make up your mind, you must approach Allah through Prayer." \\
\textbf{Source Type:} hadith \\
\textbf{Source Reference:} Paraphrase of Sahih al-Bukhari 1166 \\
\textbf{Source Text:} "If any one of you thinks of doing any job, he should offer a two-rak`ah prayer other than the obligatory ones and then say…" \\
\textbf{Source URL:} \url{https://sunnah.com/bukhari:1166} \\
\textbf{Verification Comment:} Conceptually matches the ḥadīth but the quotation is not verbatim and no reference was supplied. Treated as a paraphrase.

\end{tcolorbox}

\end{document}